\documentclass[journal]{IEEEtran}

\ifCLASSINFOpdf
\else
   \usepackage[dvips]{graphicx}
\fi

\usepackage{url}
\usepackage{graphicx}
\usepackage{xcolor}
\usepackage{cite}
\usepackage[colorlinks=true]{hyperref}
\usepackage{amsmath}
\usepackage{amssymb}
\usepackage{booktabs}
\usepackage{comment}
\usepackage{multirow}
\usepackage{amsbsy}
\usepackage{subcaption}
\usepackage{stfloats}

\begin{document}

\title{ViFi-ReID: A Two-Stream Vision-WiFi Multimodal Approach for Person Re-identification}

\author{Chen Mao, Chong Tan, \IEEEmembership{Member, IEEE}, Jingqi Hu, and Min Zheng
\thanks{}
\thanks{}
}

\markboth{IEEE Signal Processing Letters}
{Cartella \MakeLowercase{\textit{et al.}}: Unveiling the Truth: Exploring Human Gaze Patterns in Fake Images}
\maketitle

\begin{abstract}
% 行人重识别作为安防领域的重要技术，在安全检测、人员统计等方面发挥着至关重要的作用。当前重识别方案大多仅从图像中提取特征，易受外观服装、遮挡等客观条件干扰。除了摄像头，我们采用广泛存在的路由器作为感知设备，通过接收WiFi信号中的信道状态信息（CSI）来捕捉行人步态作为数据来源。我们使用双流网络分别处理视频理解和信号分析任务，并对行人的视频和WiFi数据进行多模态融合和对比学习。在现实世界中的大量实验表明，我们的方法能够挖掘异质数据的相关性，打通视觉和信号模态间的壁垒，有效扩大感知范围，并在多传感器下提升重识别的准确性。
Person re-identification(ReID), as a crucial technology in the field of security, plays a vital role in safety inspections, personnel counting, and more. Most current ReID approaches primarily extract features from images, which are easily affected by objective conditions such as clothing changes and occlusions. In addition to cameras, we leverage widely available routers as sensing devices by capturing gait information from pedestrians through the Channel State Information (CSI) in WiFi signals and contribute a multimodal dataset. We employ a two-stream network to separately process video understanding and signal analysis tasks, and conduct multi-modal fusion and contrastive learning on pedestrian video and WiFi data. Extensive experiments in real-world scenarios demonstrate that our method effectively uncovers the correlations between heterogeneous data, bridges the gap between visual and signal modalities, significantly expands the sensing range, and improves ReID accuracy across multiple sensors.
\end{abstract}


\IEEEpeerreviewmaketitle

\section{Introduction}
% 行人重新识别是安防监控领域的一项关键任务，由于相机分辨率和拍摄角度的缘故，不可能得到高质量的人脸图片来进行人脸识别，重识别就成为了重要的人员或者物体识别技术。近年来，这项任务因其在大规模监控网络中潜在的重要应用而受到学术界和工业界的广泛关注，具有重大的研究影响和实用价值。
Person ReID is a crucial task in the field of security surveillance. Due to limitations in camera resolution and shooting angles, obtaining high-quality facial images for face recognition is often not feasible. Therefore, ReID emerges as a significant technology for the identification of individuals or objects in such scenarios. In recent years, the task has attracted widespread attention from both the academic and industrial communities due to its potentially important applications in large-scale surveillance networks, demonstrating significant research impact and practical value.

% 在当前主流的基于视觉的重识别方法中，RGB图片易受光照、遮挡、行人服装的负面影响，且摄像头的部署范围有限。一些多模态的重识别方法借助红外摄像头、激光雷达等设备，可以弥补夜间识别精度低的问题，但造成较大的成本负担。一些基于WiFi信号的方法为模式识别提供了一种新思路，路由器是一种广泛存在的感知传感器，在一些不常安装摄像头的地方如室内均有分布。在无线通信中，WiFi信号受物理环境的影响可以通过信道状态信息（CSI）来描述。当行人经过时，发射器发出的信号接触到人体进行散射和反射，接收器抓取一段时间内的信号，这个信息隐式的描述了行人的步态和空间等信息，使得每个行人产生区分度。
In current mainstream vision-based ReID methods \cite{wang2018learning, 2019Bags, ye2021deep, 2020FastReID, 2021Watching, hou2021bicnet, 2021Spatial}, RGB images are susceptible to negative impacts from lighting, obstructions, and pedestrian clothing, and the deployment range of cameras is limited. Some multimodal ReID methods \cite{li2023clip, 2023Counterfactual, wu2023unsupervised, zhang2023diverse}, utilizing infrared cameras, LiDAR, and other devices, can mitigate the issue of low recognition accuracy at night but lead to a significant cost burden. Some methods \cite{li2021two, zhou2022csi, 2018Through, zhang2022tips, 2015PhaseFi, 2019Wi, 2021Multi, avola2022person, 2020Vision, deng2022gaitfi, 2019XModal, 10663544} based on WiFi signals offer a new approach to pattern recognition. Routers, as widely available sensing sensors, are distributed in places where cameras are not commonly installed. In wireless communication, the impact of the physical environment on WiFi signals can be described through CSI. As pedestrians pass by, the signals emitted by the transmitter scatter and reflect off the human body, and the receiver captures the signals over a period of time. This information implicitly describes the pedestrian's gait and posture information, enabling differentiation between individuals based on their unique characteristics.

\begin{figure}[t]
  \centering
  \includegraphics[width=\linewidth]{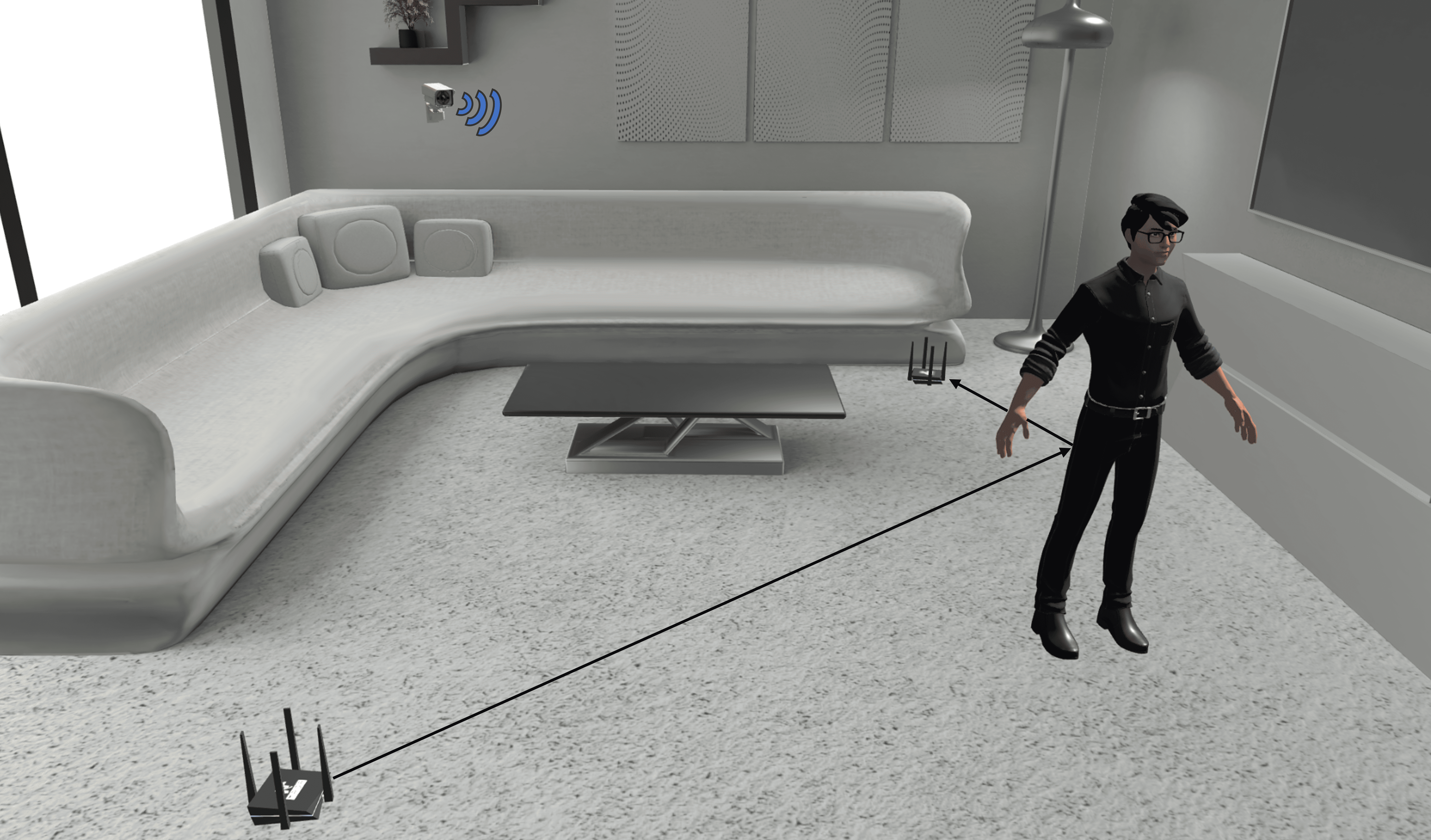}
  \caption{A pair of WiFi transceivers and a camera are used to identify the person}
  \vspace{-.35cm}
  \label{fig:figure1}
\end{figure}

% 行人的视频和WiFi时序信号均能反映行人的步态信息，视频额外包含了RGB色彩和行人动作信息，WiFi额外包含了空间定位信息，即两种模态间存在相关性和互补性。基于以上分析，提出一种基于视觉和WiFi信号的多模态的方法，它可以使用现实世界中已存在的摄像头和路由器来完成行人重识别任务。在实际场景中，如果部分区域仅存在摄像头、仅存在路由器或两者均存在，且目标行人的数据被训练过，那么可以在不同模态的场景下串联起此行人的ID，即完成针对同一行人的不同模态数据间的相互匹配检索，这保证了不同模态下重识别的连续性，而不是多个孤立的系统。在视觉和WiFi均存在的多模态场景下，我们的方法有超越单模态方法的准确性和鲁棒性。该方法在单模态和多模态的场景下均可工作，打通异质数据间的差异性，扩展重识别任务的覆盖范围。
Both pedestrian videos and WiFi temporal signals can reflect pedestrian gait information. Videos additionally include RGB color and pedestrian motion, while WiFi adds spatial positioning information, indicating that there is both correlation and complementarity between the two modalities. We propose a multimodal approach based on both vision and WiFi signals, which can utilize existing cameras and routers in the real world to accomplish the task of person ReID. In practical scenarios, if some areas have only cameras, only routers, or both, then it is possible to link this person's ID across different modal scenes. This ensures the continuity of ReID across different modalities, rather than multiple isolated systems. The method works in both single-modal and multimodal scenarios, bridging the gap between heterogeneous data and expanding the coverage of ReID tasks.

% ViFi-ReID由双流网络和特征融合网络组成。对于WiFi信号模块，我们将固定时间单位内的WiFi信号进行编码，并对这段原始编码添加时序token，实现对信号中时间信息的感知。对于视频序列模块，我们使用两个独立的Transformer编码器，由空间编码器对同一帧抽出的tokens特征提取，然后由时间编码器将不同帧的同一位置的tokens进行交互理解。在融合模块中，我们使用合并注意力机制连接两种模态的特征向量，分别将表征学习和度量学习应用于网络的最终预测和级联特征层，以生成更加全面的行人特征表示。同时，为了寻找两种模态之间的关系，我们设计基于对比学习和难例挖掘的损失函数，以实现跨模态匹配。此外，我们提出了一个具有挑战性的视觉-WiFi行人重识别数据集，包含行人的视频和WiFi数据。为以后的相关研究提供了数据支持。
For the WiFi signal module, we encode the WiFi signals within fixed time units and add temporal tokens to the raw encoding to capture the temporal information in the signals. For the video sequence module, we employ independent spatial and temporal encoder\cite{vaswani2017attention, arnab2021vivit} to extract pedestrian motion features from the video. And we use merged attention mechanism to connect the feature vectors of the two modalities. Representation learning and metric learning are applied to the network's final predictions and cascaded feature layers, generating a more comprehensive pedestrian feature representation. Additionally, to explore the relationship between the two modalities, we design a loss function based on contrastive learning and hard example mining to achieve cross-modal matching. In addition, We introduce a challenging vision-WiFi person ReID dataset, comprising both pedestrian videos and WiFi data, which provides data support for future related research.

\section{Dataset}
\begin{figure}[h]
\centering
\includegraphics[width=\linewidth]{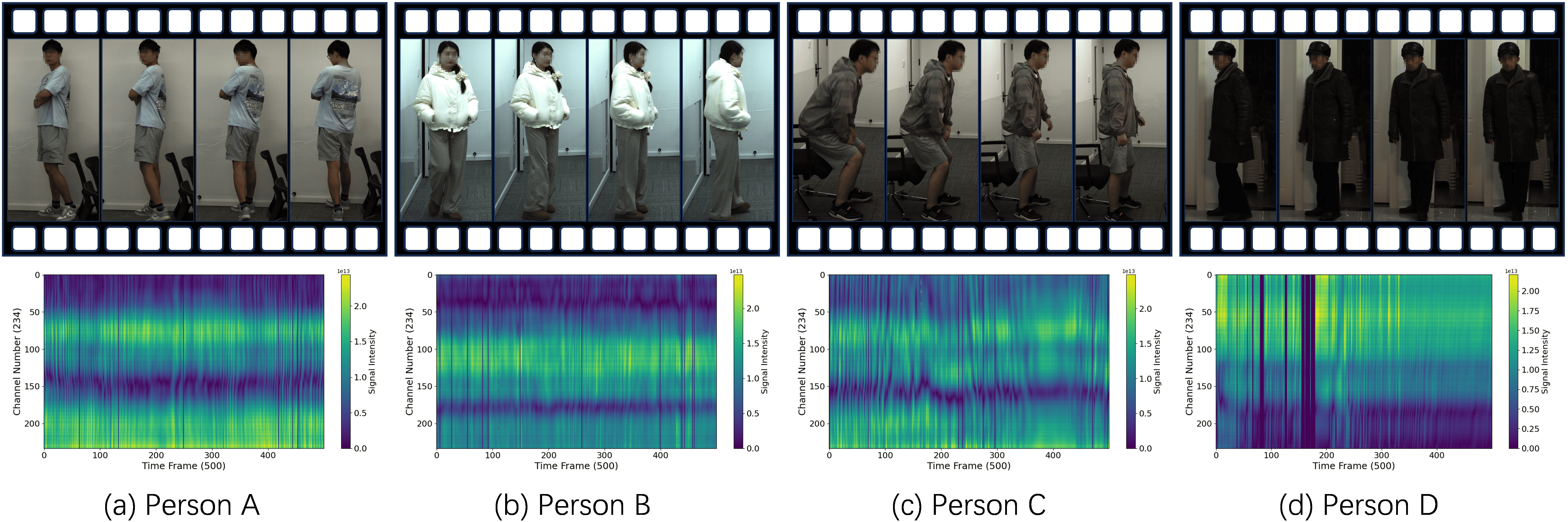}
\caption{Example of one video clip of a sample and the corresponding matrix of WiFi CSI data frame.}
\label{fig:figure2}
\vspace{-.35cm}
\end{figure}

% 现有的行人重识别数据集不包含WiFi信号信息，为了完成我们提出的模型的训练和测试任务，我们基于真实室内环境，贡献了一个名为ViFi-ReID Dataset的多模态行人重识别和检索数据集。ViFi-ReID Dataset不仅包含行人的视频序列数据，还同步采集了该场景中路由器的WiFi信号数据。基于ViFi-ReID Dataset，我们可以探索通信信号和视频数据在行人重新识别和检索方面的潜力，并验证不同方法的有效性。图3是数据集中部分视频及对应的WiFi信号可视化样例数据，下面我们简单介绍一下ViFi-ReID Dataset的采集和设置。
We contribute a multimodal person ReID and retrieval dataset named ViFi-Indoors, based on a real indoor environment. The ViFi-Indoors not only contains video sequence data of pedestrians but also synchronously collects WiFi signal data from routers in the scene. Figure \ref{fig:figure2} shows some visual examples of video and corresponding WiFi signals from the dataset. 

\begin{figure*}[bp]
\centering
\includegraphics[width=\linewidth]{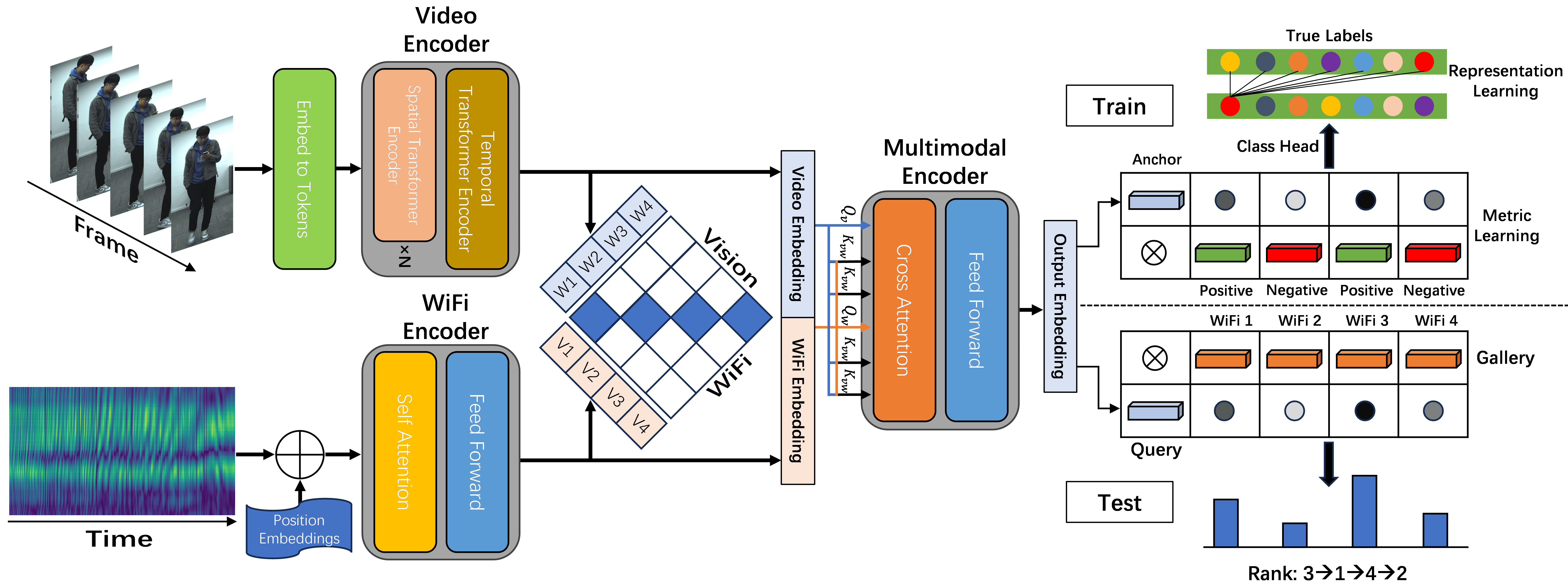}
\caption{The overall architecture diagram of ViFi-ReID.}
\label{fig:figure3}
\vspace{-.35cm}
\end{figure*}

% 视频采集。我们拍摄使用的设备为分辨率为2448*2048，型号为MV-CS00-10GC的海康威视工业相机和型号为MVL-MF0828M-8MP的定焦镜头。数据集包含三个不同的真实室内场景。为了保证数据采集的真实性，我们拍摄的行人包括男人、女人、老人和小孩等，被拍摄者存在佩戴口罩、身体部分被遮挡等情况，并存在多个行人穿着相似的衣服和单个行人在不同场景中穿着衣服不同等复杂情况，非常具有挑战性。针对每个行人，摄像头以15fps的速度连续收集5分钟视频,共4500帧。在数据预处理时，我们对采集到的视频进行3倍下采样，并通过yolov5目标检测算法对采样后的视频帧打上数据标签。我们通过检测算法生成的标签对目标边界框进行裁剪，得到仅包含行人的感兴趣区域。我们设5秒为一个视频序列单元，每个视频序列单元包含25帧连续图像，视频被分成同样大的序列。最终我们的视觉数据包含20名行人的1380视频序列，总共34500帧图像。
\textbf{Video sequence:} Our dataset contains three different real indoor scenes. To ensure the authenticity of data collection, we film a diverse range of persons including men, women, elderly, and children. The subjects being filmed are in challenging scenarios, including wearing masks, partial body obstructions, multiple persons wearing similar clothing, and a single person wearing different clothes in different scenes. Ultimately, the vision data comprises 1380 video sequences from 20 persons, totaling 34,500 frames.

% WiFi信号采集。对于发射和接受WiFi的路由器，我们均选用华硕的RT-AC86U路由器。它兼容802.11ac所定义的超高吞吐量编码，工作带宽可达80 MHz,具有242个子载波。路由器配有四根天线，支持各种发射-接收天线组合,最多 4×4 MIMO。在实际采集过程中，我们使用了234个子载波，并使用单天线发射，四天线接收，构成1*4的MIMO矩阵，WiFi信号发射的频率为每秒100次。由图d的WiFi信号可视化可以看出，WiFi数据存在丢包的情况，这些瑕疵也符合真实使用的场景。为了保持与视频信息同步，我们在采集视频的过程中同步采集WiFi信号，同样设5秒为一个WiFi信号单元，所以我们采集到一个信号单元内的WiFi信号维度为500*4*234。与视频采集相对应，最终我们的WiFi信号数据也包含20名行人的1380个信号单元。数据的同步采集与使之对齐的预处理方法为保证了两种模态数据在输入模型时有相同的颗粒度。
\textbf{WiFi signal:} We use 234 subcarriers and employ a single antenna for transmission and four antennas for reception, forming a 1$\times$4 MIMO matrix. The WiFi signals are transmitted at a frequency of 100 times per second. To maintain synchronization with the video, we concurrently collect WiFi signals during video recording, setting 5 seconds as a WiFi signal unit. Therefore, the dimension of the WiFi signal unit is $500\times4\times234$. Corresponding to the video collection, our WiFi data also includes 1380 signal units from 20 persons.

\section{Proposed Method}
% % 我们的方法使用图3的结构，构建视觉和WiFi的多模态双流网络，充分利用摄像头和路由器这种通用感知设备来获取数据，训练模型并前向推理。
% We utilize the structure presented in Figure \ref{figure2} (c), building a multimodal two-stream network for vision and WiFi. It extensively leverages common sensing devices such as cameras and routers to gather data, which is then used to train the model and perform forward inference.

% % 我们将在本节介绍提出的ViFi-ReID模型网络结构，并逐模块阐述我们的改进方法。模型框架可以分为视频特征提取网络、WiFi特征提取网络、特征融合网络和损失函数四部分。ViFi-ReID神经网络模型架构图如图4所示。
% In this section, we introduce the proposed ViFi-ReID architecture and elaborate on our improvement methods module by module. The model framework can be divided into four parts: video feature extraction network, WiFi feature extraction network, feature fusion network, and losses. 

\subsection{Video Feature Extraction}

% 我们吸收了基于Transformer架构的ViViT视频分类模型的精髓，构建两个独立Transformer编码器。空间编码器作为第一部分，仅对相同时间索引中提取的tokens间的交互作用进行建模。经过\( L_s \)层空间Transformer的处理后，我们获得了每一帧在空间维度上的特征表示\( h_i \in \mathbb{R}^{d_s} \)。接着，这些帧级特征\( h_i \)被汇聚成\( H \in \mathbb{R}^{n_t \times d} \)，并送入由\( L_t \)层Transformer构成的时间编码器中，以建模不同时间索引的tokens之间的相互作用，捕捉视频不同帧之间的动态关联。通过独立编码器的设计使模型具有更少的参数量，造成的计算复杂度为\( O((n_h \cdot n_w)^2 + n_t^2) \)，提升了计算效率。在最终输出的特征向量中，其维度被设定为batch size乘以feature dimension，其中feature dimension设置为768。

The architectural diagram of the neural network model is shown in Figure \ref{fig:figure3}. Drawing from the core principles of the ViViT video classification model based on the Transformer architecture \cite{arnab2021vivit}, we construct two separate Transformer encoders. The spatial encoder, serving as the first part, solely models the interactions between tokens extracted from identical temporal indices. After undergoing \( L_s \) layers of spatial Transformers, we acquire a spatial feature representation for each frame, \( h_i \in \mathbb{R}^{d_s} \). These frame-level features \( h_i \) are then amalgamated into \( H \in \mathbb{R}^{n_t \times d} \), which is subsequently input into a temporal encoder comprised of \( L_t \) Transformer layers. This temporal encoder is tasked with modeling interactions between tokens from disparate temporal indices, thereby capturing the dynamic correlations between various frames.
\subsection{WiFi Feature Extraction}
% 现代WiFi设备遵循 IEEE 802.11n/ac标准，在物理层使用正交频分复用，允许使用多个发射和接收天线实现多输入多输出通信。WiFi信号在路由器之间的进行传输，受物理环境影响而发生变化，WiFi中的CSI信息揭示了每个通信子载波上的延迟、幅度衰减和多径相移效应的细粒度特征。
Modern WiFi devices adhere to the IEEE 802.11n/ac standards, utilizing Orthogonal Frequency-Division Multiplexing (OFDM) at the physical layer, which allows the use of multiple transmitting and receiving antennas to achieve Multiple Input Multiple Output (MIMO) communication. WiFi signals transmitted between routers undergo changes influenced by the physical environment. The CSI in WiFi reveals the fine-grained characteristics of delay, amplitude attenuation, and multipath phase shift effects on each communication subcarrier.

\begin{figure}[t]
  \centering
  \includegraphics[width=\linewidth]{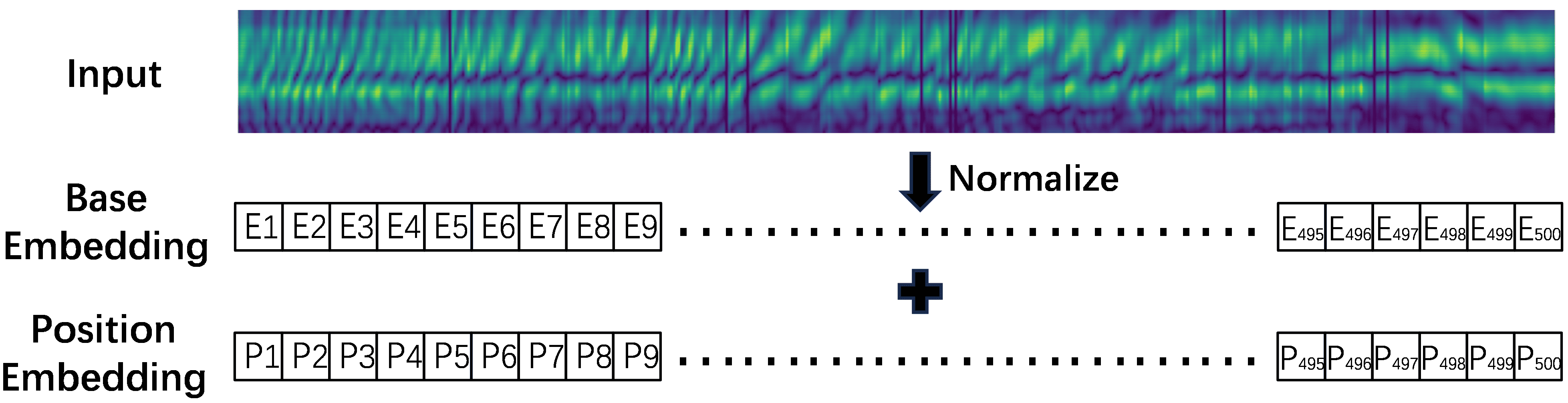}
  \caption{WiFi Data Preprocessing in WiFormer.}
  \vspace{-.35cm}
  \label{figure5}
\end{figure}

% 为了提取时序的WiFi信号的特征，我们提出一种简单可行的特征向量表示方法。我们参考了自然语言处理方法Bert的特征嵌入处理方式，由于文字不能直接输入计算机运算，需要将每个单词转换为向量。最终的向量表示由词本身的embedding、表示句子位置的embedding和字符位置的embedding相加组成。如图5所示，我们提出一种将WiFi信号进行向量表示的方法WiFormer，WiFi信号在传输过程中天然由数字组成，我们对信号的复数数据取绝对值并进行正则化后，在基础嵌入的对应时间帧上加上可学习的位置嵌入，用于表明一个CSI序列中每个时刻的位置。处理后的数据送入多层双向Transformer Encoder进行分析，使用自注意力机制来处理时序信号的长距离依赖问题。最终生成的特征向量维度和视频模块输出保持一致，这有利于两种模态特征进行对比学习和匹配检索。
To extract features from sequential WiFi signals, we propose a straightforward and practical method for representing feature vectors. As shown in Figure~\ref{figure5}, drawing inspiration from the natural language processing method BERT \cite{2018BERT}, we introduce a method named WiFormer for vector representation of WiFi signals. Given that WiFi signals are naturally composed of digits during transmission, we take the absolute value of the signal's complex data and normalize it. Then, on the base embedding of the corresponding time frame, we add learnable position embeddings to denote the location of each moment in a CSI sequence. The processed data is fed into multiple layers of bidirectional Transformer Encoders for analysis, utilizing self-attention mechanisms to address the long-distance dependencies of sequential signals.

% WiFormer中的WiFi数据预处理。将信号数据正则化处理后，基础嵌入和位置嵌入组成完整输入，并送入Transformer Encoder进行特征提取。

% 特征融合网络。
% 跨模态融合有助于实现多种异质信息的互补，多模态数据的结合可以为模型决策提供更多的信息，从而提高了最终决策结果的准确率。视频和WiFi均隐式的包含行人的行走步态信息，但各自又包含独特的特征，如视频序列包含额外的色彩信息，WiFi信号包含额外的空间感知信息。融合网络的目的是处理和关联来自多种模态的信息，实现更有效的互补。为了实现图像和WiFi模态之间的充分交互，我们通过高效的多模态交互编码器Merged attention来融合视频和WiFi嵌入，多模态交互编码器由多头交叉注意层和多层Transformer块组成。对于前一阶段从视频和WiFi双流网络中提取出的两种模态特征，我们将其拼接后送入由特征融合网络，通过跨模态的互注意力机制自适应的融合两种模态特征。最终输出特征维度仍为为batchsize*feature_dim，用于重识别任务中度量学习中的距离计算，并通过一个分类头用于实现表征学习。在推理过程中，仅需要得到倒数第二层的原始特征即可。
Cross-modal fusion facilitates the complementarity of various heterogeneous information. Both video and WiFi implicitly contain information about pedestrian walking gait, and video sequences contain additional color information, while WiFi signals provide extra spatial perception information. The goal of the fusion network is to process and correlate information from multiple modalities to achieve more effective complementarity. We employ merged attention \cite{2021An} to fuse video and WiFi embeddings. The multimodal fuse module is composed of multi-head cross-attention and multi layers of transformer encoder blocks. The features from both modalities, extracted from the previous stage's two-stream network of video and WiFi, are concatenated and fed into the feature fusion network and adaptively fused through the cross-modal attention mechanism.

\vspace{-0.35cm} 
% 使用ViFi-ReID训练集和测试集和相同的测试标准，与近些年其它的行人重识别方法进行比较，最优指标用粗体标出。
\begin{table*}[bp]
  \centering
  \caption{Comparative experiment for the comparison with other person ReID methods from recent years.}
  \label{table1}
  \begin{tabular}{c|cccc|ccccc}
    \toprule
    Method&Publication&Input Modality&BackBone&Desc. Dim.&mAP&mINP&Rank-1&Rank-5&Rank-10\\
    \midrule
    BoT \cite{2019Bags} & CVPR & Vision & ResNet-50 & 2048 & 59.65 & 48.51 & 97.78 & 99.67 & 99.89\\
    AGW \cite{ye2021deep} & TPAMI & Vision & ResNet-50 & 2048 & 66.12 & 35.90 & 97.02 & 99.27 & 99.60\\
    SBS \cite{2020FastReID} & ArXiv & Vision & ResNet-50 & 2048 & 75.86 & 55.74 & 93.76 & 94.78 & 98.89\\
    MGN \cite{wang2018learning} & ACM MM & Vision & ResNet-50 & 2048 & 71.71 & 42.42 & \textbf{99.49} & 99.89 & 99.93\\
    SimpleViTFi \cite{SimpleViTFi} & IEICE & WiFi & ViT & 768 & 70.73 & 47.08 & 92.73 & 99.09 & 99.09\\
    \midrule
    ViFi-ReID(Only Vision) & -{}-{}-{}- & Vision & ViViT & 768 & 78.63 & 57.39 & 98.18 & \textbf{100.00} & \textbf{100.00}\\
    ViFi-ReID(Only WiFi) & -{}-{}-{}- & WiFi & WiFormer & 768 & 75.49 & 59.67 & 91.82 & 98.18 & 99.09\\
    \textbf{ViFi-ReID(Full)} & -{}-{}-{}- & Vision+WiFi & ViViT+WiFormer & 768 & \textbf{79.05} & \textbf{64.92} & 96.36 & 99.09 & 99.09\\
    \bottomrule
  \end{tabular}
\end{table*}

\subsection{Representation and Metric Learning}
% 为了有效提升行人重识别的性能，我们采用了更先进的损失函数LMCL和SoftTriple，分别从表征学习和度量学习的角度优化特征空间，以增强模型的区分能力和泛化能力。

% 我们将基于表征学习的LMCL置于模型的分类头后，旨在通过最大化类内相似性和类间差异性来改善深度学习模型的表征能力。LMCL通过引入余弦距离和固定的边缘使得同类样本的特征向量更加紧密，而不同类样本的特征向量之间的距离更大。基于传统的softmax损失函数，LMCL通过对特征向量进行归一化并乘以一个缩放因子$s$来增加决策边界的间隔，具体公式为：

We place the large margin cosine loss\cite{2018CosFace} behind the classification head of the model and SoftTriple\cite{2019SoftTriple} at the feature output layer, aiming to enhance the representation capability of the model by maximizing intra-class similarity and inter-class dissimilarity and effectively learn the fine-grained structure of categories within the feature space.
\begin{equation}
L_{\text{LMCL}} = -\frac{1}{N} \sum_{i=1}^{N} \log \frac{e^{s(\cos(\theta_{y_i}) - m)}}{e^{s(\cos(\theta_{y_i}) - m)} + \sum_{j \neq y_i} e^{s \cos(\theta_j)}}
\end{equation}
% 其中，$N$是批次大小，$\theta_{y_i}$是正确类别的特征向量与权重向量之间的角度，$m$是引入的边缘，$s$是缩放因子。
where $N$ is the batch size, $\theta_{y_i}$ is the angle between the feature vector of the correct class and the weight vector, $m$ is the introduced margin, and $s$ is the scaling factor.

{
\fontsize{8pt}{10pt}\selectfont
\begin{equation}
L_{\text{SoftTriple}} = \frac{1}{N} \sum_{i=1}^{N} \log \frac{1 + \sum_{j=1}^{C} \sum_{k=1}^{K} \exp(-\sigma(d(x_i, c_{jk}) - \delta))}{\exp(\lambda) + \sum_{j \neq y_i} \sum_{k=1}^{K} \exp(-\sigma d(x_i, c_{jk}))}
\end{equation}
}
% 其中，$N$是批次大小，$C$是类别数，$K$是每个类别的中心数目，$d(x_i, c_{jk})$是样本$x_i$到其第$j$个类别的第$k$个中心$c_{jk}$的距离，$\sigma$是缩放参数，$\delta$和$\lambda$分别是控制类间和类内分离度的参数。
where $N$ is the batch size, $C$ is the number of categories, $K$ is the number of centers per category, $d(x_i, c_{jk})$ is the distance between sample $x_i$ and center $c_{jk}$ of category $j$, $\sigma$ is the scaling parameter, and $\delta$ and $\lambda$ are parameters controlling the degree of separation between and within categories, respectively.

\subsection{Contrastive Learning for Multimodality}
% 为了打通模态间的差异性，扩大重识别覆盖范围，我们通过对比学习寻找视觉和信号间的关系，已实现不同模态间的互相检索。不同于CLIP中每个样本对独立的对比学习方法，本文的多模态数据中，一个batch中每个样本可能都有很多的正对和负对，需要通过标签有监督的区分样本之间的正负关联。我们将有监督对比学习的方法用于视觉和WiFi的多模态匹配任务中，通过最小化视觉-WiFi匹配对之间的距离并最大化非匹配对之间的距离来学习特征表示，以提高特征空间中两种模态匹配对的聚合度，并增强模型对非匹配对间差异的辨别力，公式如下：
We explore the relationships between vision and signals through contrastive learning, enabling mutual retrieval across different modalities. Unlike CLIP \cite{2021Learning}, we apply a supervised contrastive learning method \cite{2020Supervised} to the multimodal matching task of vision and WiFi, learning feature representations by minimizing the distance between matching vision-WiFi pairs and maximizing the distance between non-matching pairs. This approach aims to enhance the aggregation of matching pairs in the feature space and strengthen the model's ability to differentiate between non-matching pairs. The formula is as follows:
{
\fontsize{8pt}{10pt}\selectfont
\begin{equation}
\mathcal{L}_{\text{w2v}} = -\frac{1}{N}\sum_{i=1}^{N}\frac{1}{|P(i)|}\sum_{p \in P(i)}\log\frac{\exp(\text{s}(w_i, v_p) / \tau)}{\sum_{j=1}^{N}\exp(\text{s}(w_i, v_j) / \tau)}
\end{equation}
}
\begin{equation}
\mathcal{L}_{\text{sup}} = \mathcal{L}_{\text{w2v}} + \mathcal{L}_{\text{v2w}}
\end{equation}
% 其中，N代表是一个批次中的样本数量。P(i)是批次中第i个样本的所有正样本索引集合。s(wi,vj)代表WiFi特征和视觉特征之间的相似度，即向量点积值。τ是一个温度参数，用于调节尺度。
where $N$ is the size of a batch, $P_{i}$ is the set of indices of all positive samples in the batch for the $i$-th sample. $s\left(w_{i},v_{j}\right)$ represents the similarity between the WiFi feature and the vision feature, which is the dot product of the embeddings. $\tau$ is a temperature parameter used to adjust the scale.

% 在一个batch中，匹配的视频和WiFi对均以正例的形式呈现，这使得损失函数非常容易拟合。所以对于一个batch中的每个视频，都在与它不同ID的WiFi信号中寻找一个最相似的负例，对于WiFi信号同理。这使得该任务非常具有挑战性，需要在非常相似的负例中进行分辨，大大的增强了模型的鲁棒性。具体地，我们可以定义为:
In a batch, matched video and WiFi pairs are presented as positive examples, which can make the loss quite straightforward to fit. Therefore, for each video in a batch, we look for the most similar negative example among WiFi signals with different IDs, and similarly for WiFi signals, the function can be defined as follows:

{
\fontsize{7pt}{9pt}\selectfont
\begin{equation}
\mathcal{L}_{dis} = \frac{1}{N}\sum_{i=1}^{N}[y_{i} \cdot D(v_{i},w_{i}) + (1-y_{i}) \cdot max(0, (margin - D(v_{i},w_{i})))]
\end{equation}
}
% 其中，vi和wi分别代表视频和WiFi信号的特征向量，yi是对应索引的标签，D是第i个索引的视频特征和WiFi特征间的余弦距离。
where $v_{i}$ and $w_{i}$ represent the feature embeddings of the video and WiFi signal, respectively, and $y_{i}$ is the label of the corresponding index.

% 对于视觉和WiFi之间的检索任务，我们在ViFi-ReID数据集上进行消融实验，来验证每个模块对结果指标的影响。WiFi-to-Vision表示由WiFi检索视觉信息，Vision-to-WiFi表示由视觉检索WiFi信息，VMC表示视觉-WiFi对比学习，VMD表示视觉-WiFi距离度量学习。
\begin{table*}[t]
  \centering
  \caption{Ablation study for the retrieval task between vision and WiFi.}
  \label{table2}
  \begin{tabular}{c|ccccc|ccccc}
    \toprule
    \multirow{2}{*}{Method} & \multicolumn{5}{c|}{WiFi-to-vision} & \multicolumn{5}{c}{vision-to-WiFi} \\
    & mAP & mINP & Rank-1 & Rank-5 & Rank-10 & mAP & mINP & Rank-1 & Rank-5 & Rank-10\\
    \midrule
    Baseline & 14.28 & 12.45 & 10.87 & 15.22 & 19.57 & 14.11 & 12.47 & 8.70 & 13.04 & 13.04\\
    +VWC & 77.61 & 66.62 & 78.26 & 94.57 & 98.91 & 75.01 & 66.09 & 76.09 & 91.30 & 95.65\\
    +VWC+VWD & \textbf{83.73} & \textbf{72.15} & \textbf{85.87} & \textbf{95.65} & \textbf{98.91} & \textbf{84.56} & \textbf{74.82} & \textbf{84.78} & \textbf{93.48} & \textbf{97.83}\\
    \bottomrule
  \end{tabular}
\end{table*}

\section{Experiments}
% 在这一节中，我们提供了我们在实验过程中的实施细节，评估标准等。我们在ViFi-Indoors数据集上进行单模态和多模态的对比实验，并对视觉和WiFi信号的互相检索任务进行消融实验，以验证模型在连接不同模态上的有效性。

% 我们遵循通用的行人重识别方法的指标评判方法，其主要评价方法mAP (mean Average Precision)是所有查询的平均精度（AP）的平均值，用于衡量模型检索到相关行人的准确性和完整性。每个查询的AP是根据模型返回的相关行人排序的准确性计算的。mAP值越高，表示模型的性能越好。mINP (mean Inverse Negative Penalty)用于衡量模型检索相关行人时的稳健性。mINP考虑了检索列表中不相关行人的位置，对靠后的不相关行人给予更小的惩罚。这使得mINP在某些情况下比mAP更能反映模型的实际性能。Rank-N指的是在测试集中，正确的匹配出现在模型返回的前N个结果中的频率。例如，Rank-1意味着模型返回的第一个结果就是正确匹配的频率。Rank-N越高，表示模型的性能越好。ROC (Receiver Operating Characteristic) 曲线是通过改变匹配阈值来绘制的，它显示了模型在不同阈值下的真正例率（TPR）和假正例率（FPR）。ROC曲线下面积越接近1，表示模型的性能越好。

% 我们的方法是在pytorch深度学习框架中实现的，在单个型号为4090，显存24G的设备上进行训练。对于视频序列，每一帧图片的大小被调整为256×128，并会通过随机水平翻转、填充、裁剪和擦除来实现数据增强，每15帧连续图像组成一个视频序列，一个batch的大小是16。对于WiFi信号，我们将接收天线的维度4和信号子载波通道数234平铺成维度936，500个时刻的信号组成一个单元，和视频保持相同的batchsize。视频和WiFi特征提取模块分别包含8个和4个Transformer encoder，两模块各自生成的嵌入维度和模态融合后的嵌入维度均为768，比大部分ReID方法输出维度小，可以加快推理时匹配速度。我们使用Adam优化器，花费10个epoch将学习率从3.5×10−6 线性增加到1×10−4，然后在第40和90个epoch时学习率衰减到原来的0.1倍，模型共训练了120个epoch。对于ReID和不同模态间检索任务中样本匹配程度的计算，我们采用余弦距离最近邻作为衡量query和gallery间是否匹配的标准，距离越近gallery中对应样本的排名越靠前。
\subsection{Implementation Details}
Our method is implemented within the PyTorch deep learning framework and trained on a single device equipped with a 4090 model GPU and 24G of VRAM. We use the Adam optimizer, spending 10 epochs to linearly increase the learning rate from 3.5$\times$$10^{-6}$ to 1$\times$$10^{-4}$, then reduce the learning rate to 0.1 of its original at the 40-th and 90-th epochs, with the model being trained for a total of 120 epochs.
\subsection{Comparison with State-of-the-Art methods}
% 我们使用在本文提出的ViFi-ReID数据集上对本模型和近些年其它优秀的模型进行实验和测试。我们在表格1上展示了各模型在ViFi-ReID测试集上的结果，图中列出了ViFi-ReID仅有视觉信息的模型，仅有WiFi信号的模型和全模型。为了公平的比较，如果单模态方法使用基于图片的重识别模型，那么在训练和测试阶段将与ViFi-ReID使用相同的图片数据量，推理过程中使用的query和gallery也是图像。可以看出基于纯视觉的方法具有更高的Rank-1指标，我们发现纯视觉的方法中，模型对RGB色彩敏感度较高，所以行人视觉信息的query会比较轻松的在gallery中检索到该行人同样服装的视觉信息。但是同一行人不同服装的视觉信息会得到较低的相似度得分，这也导致纯视觉方法的在更综合的指标如mAP和mINP无法取得很好的效果。与其它方法相比，我们全模型在测试集上的mAP、mINP指标比当前单模态的重识别方法优秀，我们的方法在仅用一个模态的情况下依然具有竞争力，多模态全模型有着更高鲁棒性和准确度。
We conduct experiments on the ViFi-Indoors proposed in this letter, comparing our model with the state-of-the-art models. The results on the ViFi-Indoors test set for various models are displayed in Table \ref{table1}, which includes models with only visual information, only WiFi signals, and the full model utilizing both modalities of ViFi-ReID. It is evident that methods based purely on vision achieve higher Rank-N metrics. We observe that in purely visual approaches, the model is highly sensitive to RGB colors, which allows queries with a person's visual information to easily retrieve the same person's visual information with identical clothing in the gallery. However, visual information of the same person in different clothing receives lower similarity scores. Compared to other methods, our full model demonstrates superior performance in the mAP and mINP metrics, surpassing current single-modality ReID methods. Even when utilizing only one modality, our method remains competitive, and the full model exhibits higher robustness and accuracy, underscoring its effectiveness in the field of person ReID. 

\begin{figure}[htbp]
  \centering
  \includegraphics[width=\linewidth]{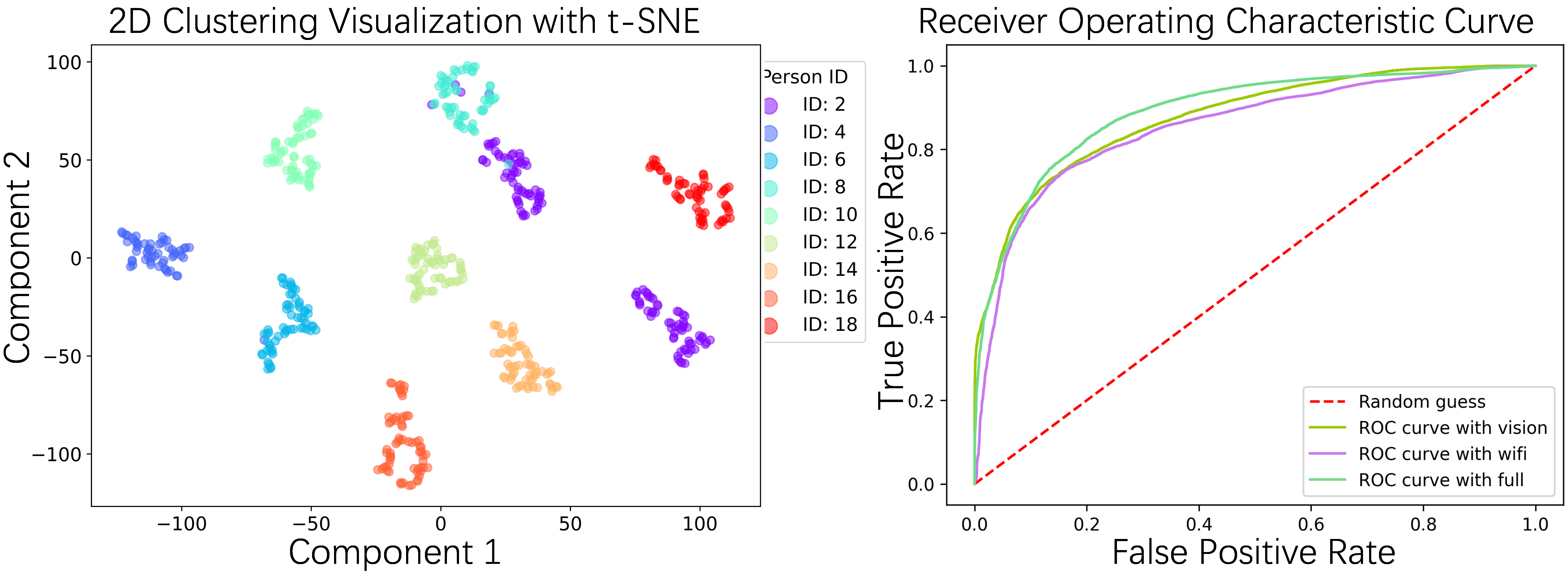}
  \caption{Visualizations of pedestrian feature clustering in two-dimensional space and ROC curve.}
  \vspace{-.35cm}
  \label{figure6}
\end{figure}

% 由t-SNE聚类算法将模型输出的行人多模态特征降维到二维空间的效果图和ViFi-ReID的仅单模态和多模态方法在测试集上ROC曲线如图所示，可以看出本方法能够在高纬空间中相同行人特征显著的进行聚类，且多模态模型能够更加准确的区分正类和负类的能力，具备优秀的召回率和精确率。
Figure~\ref{figure6} is shown that our method can significantly cluster the same pedestrian features in high-dimensional space, and the multimodal model can more accurately distinguish between positive and negative classes, exhibiting excellent recall and precision rates.

\subsection{Ablation study}
% 对于重识别任务，表1的下方可以体现出，随着模态的增多，ViFi-ReID的多模态方法优于该模型所包含的单模态方法，表明视觉和WiFi可以实现异质信息互补，提供比单一模态更丰富、更全面的信息表达。
The bottom three rows of Table \ref{table1} demonstrate that with the addition of modalities, ViFi-ReID's multimodal method indicates that vision and WiFi can complement each other with heterogeneous information, providing richer and more comprehensive expression of information than single-modality.

% 对于跨模态的检索任务，为了探索不同模块对整体性能的影响，我们进行消融实验。根据重要指标mAP和mINP, Rank-N进行评估。我们将仅进行ReID任务学习，而不刻意学习视觉和WiFi之间对应关系的模型作为baseline，逐步加入我们的方法。由表2可以看出，我们提出的方法能够显著提高不同模态间检索的精确度和召回率，WiFi检索视觉和视觉检索WiFi分别达到了85.87%和84.78%的Rank-1准确率。这表明同一行人的不同模态数据间可以实现较高准确度的检索，以显著在不同传感器下提高ReID的感知范围。我们的方法对单模态与多模态特征间的互相检索同样适用，我们将在附录中展现这部分的实验结果。
We use a baseline model that only undertakes ReID task learning without specifically learning the correspondence between vision and WiFi, and progressively incorporate our methods. As shown in Table \ref{table2}, our proposed approach significantly improves the precision and recall of cross-modal retrieval, achieving Rank-1 accuracies of 83.73$\%$ for WiFi-to-vision and 84.56$\%$ for vision-to-WiFi. This indicates that high-accuracy retrieval can be achieved between different modal data for the same person, significantly enhancing the perception range of ReID under different sensors.

% 一个由WiFi检索视觉信息，查询的前10个检索结果的图示。每个图像表示一个视频序列，匹配和不匹配的结果分别用绿色和红色矩形标记
\begin{figure}[htbp]
  \centering
  \includegraphics[width=\linewidth]{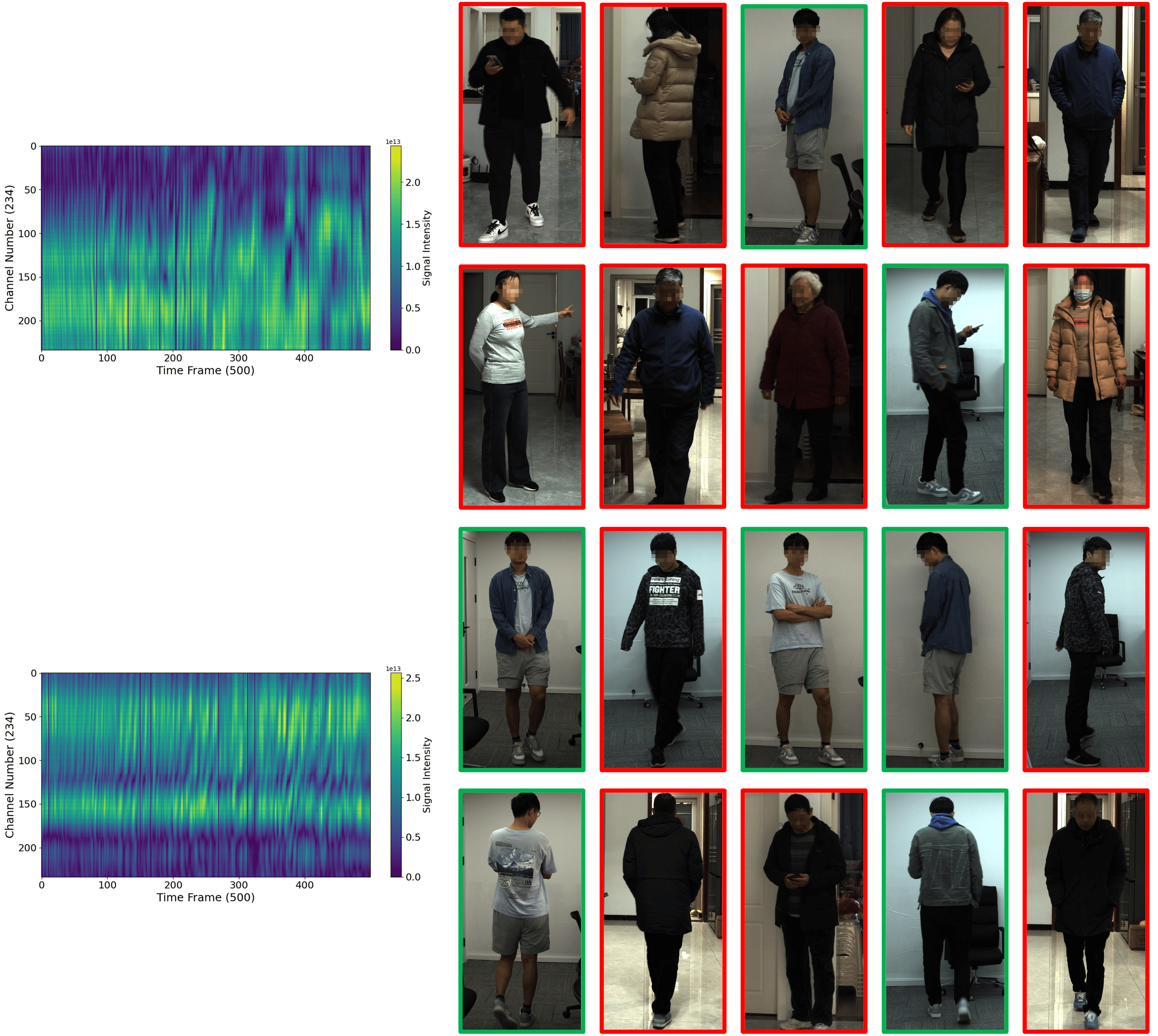}
  \caption{Comparison of top-10 retrieved results on ViFi-Indoors between baseline (the first illustration) and our final method (the second illustration) for each WiFi query.}
  \vspace{-.35cm}
  \label{figure7}
\end{figure}

% 下面是Baseline和我们最终的方法的前10个检索结果。如图所示，加入本文方法后的模型取得了更加准确的检索结果，而baseline很难获取正确的多模态间的对应关系。这主要归功于我们添加的对比学习和基于难例挖掘的度量学习，它们充分的将同一行人的两种模态的特征映射到同一个特征空间，以增强互相检索的能力。
Figure \ref{figure7} shows the top 10 retrieval results of the baseline and our final method, our method achieves more accurate retrieval results, matching and non-matching results marked with green and red rectangles, respectively.

\section{Conclusion}
% 在本文中，我们提出了一种利用视觉和WiFi的行人重识别方法，缓解了单视觉模态方法鲁棒性不强的弊端，并通过打通异质数据的壁垒扩大任务的适用范围。该方法通过双流网络进行模态分析，通过对比学习关联起模态间的对应关系，以完成视频和WiFi间的检索任务。不同模态特征通过merged attention进行多模态融合，由先进的表征和度量学习方法对融合后的综合特征进行聚类。我们在真实生活场景中采集了一个包含行人视频和WiFi信号的数据集，用于对本文提出的模型和其它优秀重识别模型进行训练和测试，并进行结果比较。根据实验结果，ViFi-ReID可以实现79.05%的mAP和96.36%的Rank-1准确率，明显优于其他基于WiFi或摄像头的单模态行人重识别方法。并在视频-WiFi对应检索任务上取得了不错的精度，证明我们的方法确实可以连接视觉和信号模态，适用于更复杂的感知场景。
We propose a person ReID method that utilizes both vision and WiFi, designed to address the limitations in robustness of single visual modal approaches and to expand the applicability of the task by bridging the gap between heterogeneous data types. According to our experimental results, ViFi-ReID outperforms other single-modal methods based on either WiFi or cameras. Additionally, it achieves impressive accuracy in vision-WiFi retrieval tasks, demonstrating that our method effectively bridges visual and signal modalities, making it suitable for more complex perception scenarios.

\bibliographystyle{IEEEtran}
\bibliography{bibliography}
\end{document}